\documentclass[sigconf, nonacm]{acmart}

\usepackage{microtype}
\usepackage{graphicx}
\usepackage{mdframed}
\usepackage{tcolorbox}
\usepackage{enumitem}
\usepackage{booktabs}
\usepackage{amsmath}
\usepackage{listings}
\usepackage{multicol}
\usepackage{array}
\usepackage{hyperref}

\tcbuselibrary{listingsutf8}
\tcbset{listing engine=listings}
\newtcblisting{promptbox}{
  colback=blue!5!white, colframe=blue!75!black,
  listing only, listing options={basicstyle=\ttfamily\small,breaklines=true},
  left=1mm, top=1mm, bottom=1mm, right=1mm,  
  boxrule=0.3mm,  
  width=\textwidth,  
}

\newcommand{\name}{HierTOD}

\newcommand\vldbyear{2025}
\newcommand\vldbworkshop{Data Science with Human-in-the-loop}
\newcommand\vldbauthors{\authors}
\newcommand\vldbtitle{\shorttitle} 
\newcommand\vldbavailabilityurl{URL_TO_YOUR_ARTIFACTS}
\newcommand\vldbpagestyle{plain} 

\begin{document}
\title{HierTOD: A Task-Oriented Dialogue System Driven by Hierarchical Goals}

\author{Lingbo Mo}
\affiliation{%
  \institution{The Ohio State University}
}
\email{mo.169@osu.edu}

\author{Shun Jiang}
\affiliation{%
  \institution{Adobe Inc.}
}
\email{shunj@adobe.com}

\author{Akash Maharaj}
\affiliation{%
  \institution{Adobe Inc.}
}
\email{maharaj@adobe.com}

\author{Bernard Hishamunda}
\affiliation{%
  \institution{Adobe Inc.}
}
\email{hishamun@adobe.com}

\author{Yunyao Li}
\affiliation{%
  \institution{Adobe Inc.}
}
\email{yunyaol@adobe.com}

\begin{abstract}
Task-Oriented Dialogue (TOD) systems assist users in completing tasks through natural language interactions, often relying on a single-layered workflow structure for slot-filling in public tasks, such as hotel bookings. However, in enterprise environments, which involve rich domain-specific knowledge, TOD systems face challenges due to task complexity and the lack of standardized documentation. In this work, we introduce \name{}, an enterprise TOD system driven by \textit{hierarchical} goals that can support \textit{composite} workflows. By focusing on goal-driven interactions, our system serves a more proactive role, facilitating mixed-initiative dialogue and improving task completion. Equipped with components for natural language understanding, composite goal retriever, dialogue management, and response generation, backed by a well-organized data service with domain knowledge base and retrieval engine, \name{} delivers efficient task assistance as judged by human evaluators. Furthermore, our system implementation unifies two TOD paradigms: slot-filling for information collection and step-by-step guidance for task execution. Our user study demonstrates the effectiveness and helpfulness of \name{} in performing both paradigms.
\end{abstract}

\maketitle

\pagestyle{\vldbpagestyle}
\begingroup\small\noindent\raggedright\textbf{VLDB Workshop Reference Format:}\\
\vldbauthors. \vldbtitle. VLDB \vldbyear\ Workshop: \vldbworkshop.\\ 
\endgroup
\begingroup
\renewcommand\thefootnote{}\footnote{\noindent
This work is licensed under the Creative Commons BY-NC-ND 4.0 International License. Visit \url{https://creativecommons.org/licenses/by-nc-nd/4.0/} to view a copy of this license. For any use beyond those covered by this license, obtain permission by emailing \href{mailto:info@vldb.org}{info@vldb.org}. Copyright is held by the owner/author(s). Publication rights licensed to the VLDB Endowment. \\
\raggedright Proceedings of the VLDB Endowment. 
ISSN 2150-8097. \\
}\addtocounter{footnote}{-1}\endgroup

\ifdefempty{\vldbavailabilityurl}{}{
\vspace{.3cm}
\begingroup\small\noindent\raggedright\textbf{VLDB Workshop Artifact Availability:}\\
The demo of \name{} system is available at \url{https://www.youtube.com/shorts/_v8f3x5IKEw}.
\endgroup
}

\section{Introduction}

Task-oriented dialogue (TOD) systems aim to help users accomplish specific goals by executing tasks through natural language interactions. Significant advancements have been made in developing systems that support public domain tasks~\citep{andreas2020task, peng2021soloist, su2022multi}, such as booking hotels or reserving restaurants. These tasks typically feature straightforward, single-layered workflows with well-defined steps, intents, and information extraction requirements. However, in enterprise environments rich with domain-specific knowledge, TOD systems face unique challenges due to the complexity and lack of standardized documentation of tasks.

\begin{figure}[t]
  \centering
  \includegraphics[width=\linewidth]{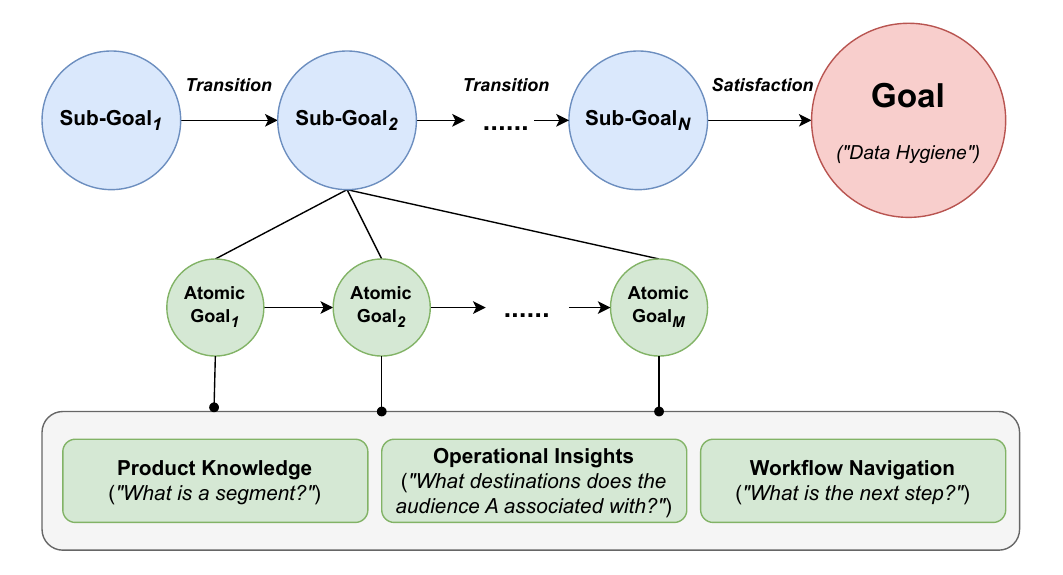}
  \caption{Composite workflows to execute tasks for hierarchical goal completion. A high-level goal consists of multiple sub-goals that transition between each other as the conversation progresses. Each sub-goal can include atomic goals such as product knowledge QA, operational insights QA, or navigation instructions.
}
  \label{goal_transition}
\end{figure}

Enterprise tasks often involve multi-layered structures composed of numerous interconnected subtasks, as illustrated in Figure~\ref{goal_transition}. These tasks are rarely formally documented, and their execution heavily relies on the implicit knowledge of human experts. When users interact with a dialogue system in such contexts, their utterances can pertain only to atomic tasks defined at the leaf nodes of a complex task hierarchy. The overarching structure and sequence of these tasks remain internalized within the user's expertise, making it difficult for dialogue systems to fully comprehend and assist effectively.

Consider, for instance, an enterprise marketing platform equipped with audience segmentation functionality. Users within an organization may independently create many client profile segments for various projects. Over time, this practice can result in the proliferation of duplicate segments, increasing platform resource costs and reducing business efficiency. To mitigate this, users perform data hygiene processes to clean up and consolidate these duplicate segments. However, the formal steps for this process are typically undocumented, and practices vary across different organizations. With guidance from subject matter experts, the complex task of identifying and removing duplicate segments can be distilled into a high-level goal comprising four sequential steps:

\vspace{5pt}
\begin{mdframed}[backgroundcolor=cyan!10] 
\small 
\noindent
\textbf{Goal: Data Hygiene for Audience Segments}

\begin{itemize}[leftmargin=*]
\setlength\itemsep{0em}  
\item Step 1: Detect duplicate segments by definition or outcome. 
\item Step 2: List segment references by relevant business entities. \item Step 3: Remove or unlink non-essential segment references and relink to essential ones when necessary.
\item Step 4: Delete non-essential segments. 
\end{itemize}
\end{mdframed}
\vspace{5pt}

In practice, different users may prioritize certain business entities over others and follow various paths through this task hierarchy when interacting with the dialogue system. Their intents correspond to atomic actions represented as nodes within a task graph, and the sequences connecting these nodes can vary significantly. A dialogue system with a deep understanding of these high-level business goals and the complex structure of such tasks can significantly enhance user experience. It can improve query comprehension, disambiguate user intents more effectively, proactively suggest relevant goals, and provide personalized responses that align with individual user needs.

To this end, we introduce \name{}, an enterprise TOD system driven by hierarchical goals to facilitate the generation of more proactive and effective dialogues with users. For example, when a user inquires about detecting duplicate segments (as described in the first step of the previous case), \name{} can proactively suggest transitioning to the high-level goal of conducting data hygiene for audience segments. Our system comprises several modules, including natural language understanding, composite goal retriever, dialogue management, and response generation, supported by a well-organized data service with a domain knowledge base and retrieval engine.

Furthermore, existing TOD systems typically follow one of two paradigms, which are often developed separately. The first is slot-filling for information collection~\citep{yang2021ubar,hu2022context,hudevcek2023large}, where users provide details and direct the system to perform specific tasks, such as making reservations. The second paradigm is step-by-step guidance, designed to assist users in executing real-world tasks by providing accurate information and step-by-step instructions. For instance, Amazon's Alexa Prize Taskbot~\citep{gottardi2022alexa, agichtein2023advancing} helps users complete tasks such as cooking a dish or following a DIY tutorial, guiding them through the process with detailed instructions~\citep{mo2023roll}. In this work, \name{} serves a unified framework that supports both paradigms, delivering comprehensive and efficient task assistance. We implement \name{} and integrate it into a prototype version of the AI Assistant~\citep{maharaj2024evaluation} within the Adobe Experience Platform\footnote{\href{https://experienceleague.adobe.com/en/docs/experience-platform/ai-assistant/home}{https://experienceleague.adobe.com/en/docs/experience-platform/ai-assistant/home}}.

Our contributions include: (1) Developing a TOD system to support composite workflows in enterprise environments. (2) Introducing a goal-driven approach to dialogues, making the system more proactive and enabling mixed-initiative interactions for improved task completion. (3) Implementing a unified framework that integrates two TOD paradigms: slot-filling for information collection and step-by-step guidance for task execution. (4) Conducting a user study to verify the effectiveness and helpfulness of our dialogue system.

\section{Related Work}


Task-Oriented Dialogue (TOD) systems are designed to help users accomplish goals through conversational interactions. TOD systems are typically categorized into pipeline-based and end-to-end approaches. Pipeline-based methods decompose the dialogue process into modular components, including natural language understanding, dialogue state tracking, dialogue policy, and natural language generation, with each module handling a distinct subtask~\citep{peng2018deep, chen2019semantically}. In contrast, end-to-end methods integrate these components into a unified model trained jointly to optimize response generation~\citep{yang2021ubar, hosseini2020simple}.

The emergence of large language models (LLMs) pretrained on open-domain data has significantly improved performance across various TOD modules~\citep{li2022controllable, zhao2023anytod, xu2024rethinking}. These models can generalize to unseen tasks with minimal supervision through zero-shot or few-shot learning, reducing reliance on extensive annotated datasets. However, while LLMs excel in generic domains, they often struggle in enterprise settings, which involve closed-source knowledge and complex task structures. To address this, HierTOD is proposed as an enterprise TOD system driven by hierarchical goals. It is designed to support composite workflows by integrating enterprise-specific knowledge bases and a composite goal retriever.

Furthermore, traditional TOD systems are largely user-driven, where the user provides structured input to complete tasks such as hotel reservations or flight bookings~\citep{andreas2020task, peng2021soloist, su2022multi}. In contrast, systems developed for the Alexa Prize TaskBot Challenge\citep{gottardi2022alexa, agichtein2023advancing} focus on interactive task guidance, helping users complete real-world activities like cooking or DIY projects by delivering step-by-step instructions\citep{mo2023roll}. HierTOD aims to unify these two paradigms, including both slot-filling for information collection and instructional guidance for task execution within one single framework, enabling more flexible and robust dialogue support across a wide range of user goals.

\section{System Design}

\begin{figure*}[t]
  \centering
  \includegraphics[width=\linewidth]{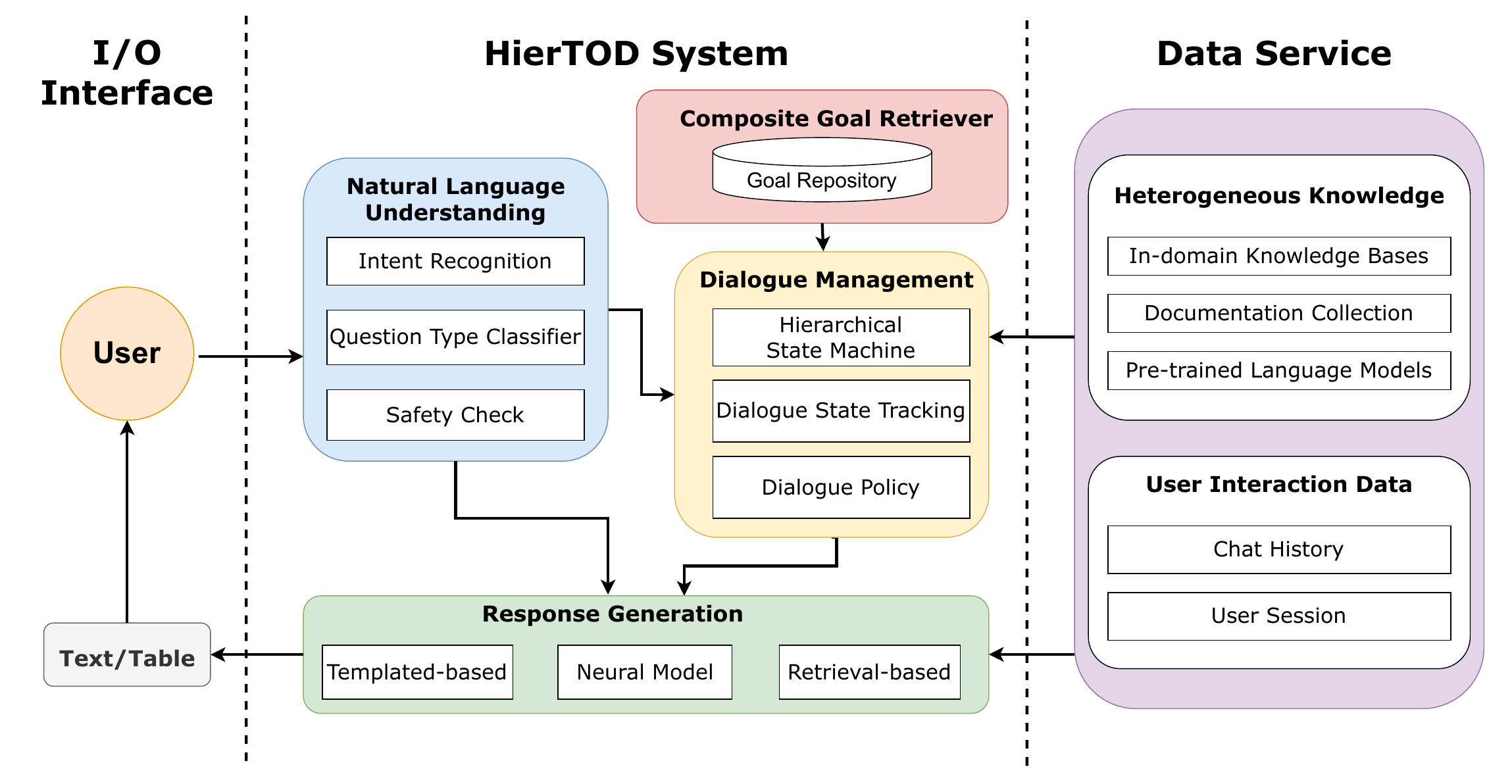}
  \caption{Architecture Overview. \name{} consists of four major components: Natural Language Understanding, Composite Goal Retriever, Dialogue Management, and Response Generation. These components are supported by a data service that integrates heterogeneous knowledge and has access to user interaction data.}
  \label{architecture}
\end{figure*}

\subsection{System Overview}
\label{sec:overview}

\name{} follows a canonical pipeline approach for TOD systems. The system consists of Natural Language Understanding (NLU), Dialogue Management (DM), and Response Generation (RG). A key feature of our system is the introduction of a Composite Goal Retriever (CGR) which  builds a goal repository to define and store workflows for goal completion.
Upon receiving user input, the NLU module (Section~\ref{sec:nlu}) preprocesses the utterance to determine the user's intent. The CGR module (Section~\ref{sec:goal_repo}) then matches the user query against the workflows in the goal repository to identify the appropriate dialogue paradigm, whether slot-filling for information collection or step-by-step guidance for task execution. Once a goal is initiated, the DM module (Section~\ref{sec:dm}), which is designed as a hierarchical finite state machine, controls the dialogue flow, handles exceptions, and guides the conversation toward task completion. The RG module (Section~\ref{sec:rg}) then generates responses or answers user queries based on intent and dialogue history. It is supported by a well-organized data service, including a domain-specific knowledge base and retrieval engine, which connects to various sources to provide optimal user assistance. We describe the details for each module in the following sections.

\subsection{Natural Language Understanding}
\label{sec:nlu}
\name{} employs a robust NLU module which combines the strengths of both trained models and rule-based approaches. The key component is Intent Recognition, where we organize multiple intents into three categories to accommodate different user initiatives, as detailed in Table~\ref{intent_cat}. To categorize different types of queries, we train a three-way classification model to classify user queries as either product knowledge, operational insights, or out-of-scope questions. For other intents, we utilize heuristics and keyword lists for recognition.

\begin{table}[ht]
\centering
\small
\renewcommand{\arraystretch}{1.5}
\begin{tabular}{m{1.5cm}|m{5.5cm}}
\hline
\textbf{Category} & \centering \textbf{Description} \tabularnewline
\hline
Sentiment & The user can confirm or reject the system's response in each turn, resulting in two labels: \textbf{Acknowledge} and \textbf{Negation}, indicating the polarity of the user's utterance. \\
\hline
Commands & The user can guide the conversation using commands such as: \textbf{Goal Trigger}, \textbf{Navigation} (e.g., moving to previous or next steps), \textbf{Task Completion}, and \textbf{Stop} to terminate the conversation at any point. \\
\hline
Utilities & The \textbf{Question} intent is used to capture various types of queries, including those related to product knowledge, operational insights, or out-of-scope questions. \\
\hline
\end{tabular}
\caption{Categories of detailed intents to support diverse user initiatives.}
\label{intent_cat}
\end{table}

\subsection{Composite Goal Retriever}
\label{sec:goal_repo}

We define the workflow for hierarchical goal completion in our dialogue system, as illustrated in Figure~\ref{goal_transition}. Specifically, a \textit{high-level goal} consists of multiple \textit{sub-goals} that can transition from one to another based on how the conversation proceeds. Each sub-goal may involve various types of QA interactions and navigation instructions. To this end, we establish and maintain a repository that defines workflows for various goals using a YAML structure. Each workflow includes a high-level goal description, followed by a series of steps to achieve that goal. Each step contains a summary, along with a detailed explanation of the actions required for completion. Based on the user’s query, we employ a CGR module that matches the query to the goals defined in the repository, using both lexical and semantic matching, to determine whether a goal is triggered, and if so, which goal is activated. If no goal is matched, the system defaults to the standard dialogue mode, focusing on question answering and general conversation.

Additionally, the goal repository accommodates both dialogue paradigms: slot-filling for information collection and step-by-step guidance for task execution. To distinguish between the two paradigms, we introduce the keyword `slots' to specify the information required for task completion. For example, when the user would like to create a ticket on the enterprise platform for assistance, the necessary slots might include `ticket title', `detailed ticket description', `priority', and `phone number'. This repository can be easily expanded or modified by domain knowledge experts to support new goals.

\subsection{Dialogue Management}
\label{sec:dm}

We design a hierarchical finite state machine for the dialogue management component, consisting of two phases: \textit{Goal Pending} and \textit{Goal Execution}. Each phase contains multiple fine-grained dialogue states. In the Goal Pending phase, users interact with \name{} by issuing a query. If the query matches a high-level goal in the Goal Repository (e.g., ``How to perform data hygiene to delete duplicate audience segments?''), the system provides an overview of the task, summarizing the upcoming steps, and then enters into the Goal Execution phase to guide the user step by step.

If the query matches a sub-goal (e.g., ``List the duplicate segments for me.''), the system provides an answer to the sub-goal and proactively asks a clarifying question for goal transition, such as, ``Would you like to delete the duplicate segments?'' If the user agrees to pursue the proposed high-level goal, they move to the Goal Execution phase until task completion. The initial step related to the matched sub-goal is skipped, as it has already been addressed. When the user query does not trigger any goal from the repository, the system utilizes the QA module to provide an appropriate response. Additionally, during the Goal Execution phase, the QA module remains available to support the user with relevant questions.

\paragraph{Dialogue State Tracking.}
For the step-by-step guidance dialogue paradigm, we use the hierarchical state machine mentioned above to keep track of the states throughout the conversation. In order to further support the slot-filling paradigm, we employ zero-shot learning using GPT-3.5~\citep{chatgpt} to perform dialogue state tracking (DST). In the designed prompt, general instructions are provided to capture values for the required slots based on both the dialogue history  and the current user utterance (see Appendix~\ref{app:prompt_dst} for the exact prompt). The updated belief state is then utilized in the subsequent response generation component described in Section~\ref{sec:rg}.

\paragraph{Dialogue Policy.}
The dialogue policy takes inputs from the NLU and CGR modules, synchronizes with the hierarchical state machine to query and update the task step state for step-by-step guidance dialogues. It also interacts with the DST to manage and update the task-specific belief state for slot-filling dialogues. The policy then drives the RG module to generate an appropriate response to the user's utterance, which will be described in the next section.

\begin{figure}[t]
  \centering
  \includegraphics[width=0.5\textwidth]{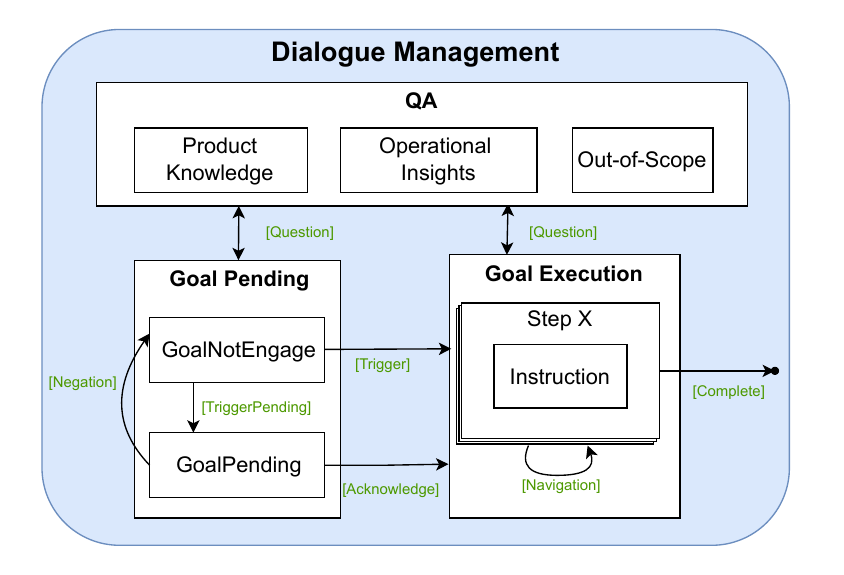}
  \caption{Dialogue Management Diagram consisting of two phases: Goal Pending and Goal Execution. White boxes represent dialogue states, while green text in brackets indicates user intent.}
  \label{dm}
\end{figure}

\subsection{Response Generation}
\label{sec:rg}

Our response generation module blends both infilling-based methods and neural models. For the step-by-step guidance dialogue paradigm, we use handcrafted conditional rules to organize curated templates and define their composition strategy according to the high-level states in our hierarchical state machine. For the slot-filling dialogue paradigm, we utilize the belief state from the DST module and make an LLM call to GPT-3.5. This can generate either a question to request missing slots or a summary to conclude the task when all required slots have been filled. The designed prompt includes the task description, belief state with filled and missing slots, dialogue history, and current user utterance, instructing the model to generate an appropriate response (see Appendix~\ref{app:prompt_rg} for the exact prompt).

In parallel, we develop a QA module to provide answers when users have questions during goal execution. As mentioned earlier, we first implement a routing model—a three-way classifier—that categorizes user questions into product knowledge, operational insights, or out-of-scope inquiries. For out-of-scope questions, we provide a predefined template response. We follow the AI Assistant~\citep{maharaj2024evaluation} within the Adobe Experience Platform for the QA module implementation. The handling of the first two types of questions is detailed in the following parts.

\textbf{Product Knowledge QA.} Product knowledge refers to concepts and topics grounded in the product documentation. Product knowledge questions can be further specified into the following sub-groups, including pointed learning, open discovery, and troubleshooting. The Product Knowledge QA component identifies the relevant documentation to answer a given question, retrieves the appropriate content, generates a response based on the retrieved information, determines proper source citations, and verifies that the responses are well-grounded.

\textbf{Operational Insight QA.} Operational insights refer to the information about metadata objects, such as attributes, audiences, dataflows, datasets, destinations, journeys, schemas, and sources including counts, lookups, and lineage impact. For example: ``How many datasets do I have?'' or ``How many schema attributes have never been used?'' The Operational Insights QA component translates user questions into SQL queries against the underlying operational data specific to customers, executes the queries to generate accurate responses, and provides comprehensive explanations for both the query and the resulting answer. These explanations include: (1) query interpretation, offering a concise summary of the actions taken to address the user's question; (2) a step-by-step breakdown of how the query is processed; and (3) inline SQL comments, allowing users with strong technical backgrounds to verify the SQL logic. These elements not only enhance transparency but also empower users to gain deeper insights into the underlying data processes.

\subsection{NL2Goal Translator}

To further simplify the goal workflow creation process, we develop an automatic NL2Goal translator powered by in-context learning using GPT-3.5. This module takes a manually crafted goal description in natural language and translates it into a structured goal definition, which is then stored in the composite goal repository in YAML format. By automating the creation of composite goals, this approach enhances the flexibility to expand and modify the goal repository, making it more feasible to adopt our goal-driven dialogue generation system in knowledge-rich domains at enterprise scale. The exact prompt for this module can be found in Appendix~\ref{app:prompt_translator}.

\section{User Study}

\begin{figure}[t]
  \centering
  \includegraphics[width=0.5\textwidth]{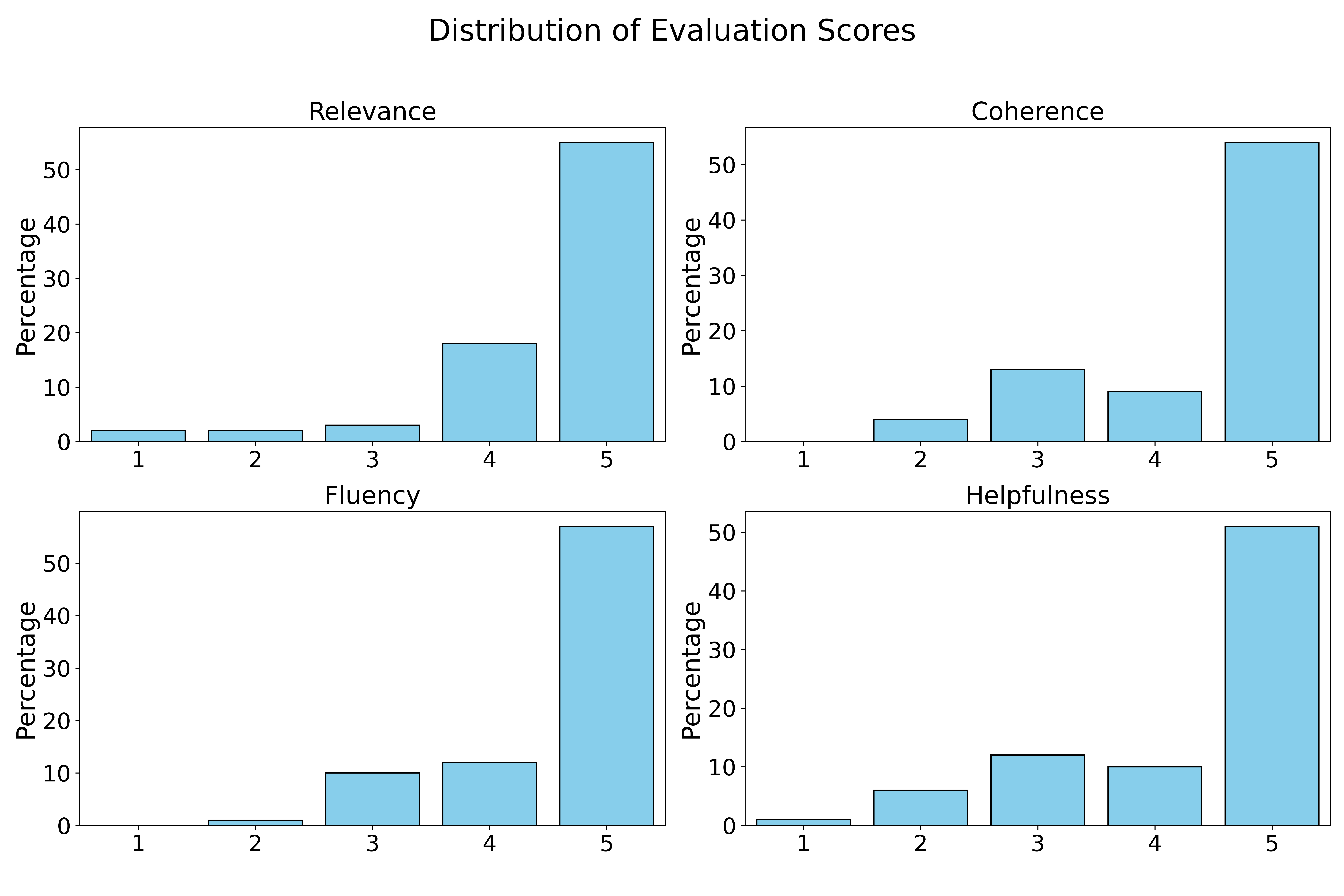}
  \caption{Distribution of evaluation scores across four aspects in the user study.}
  \label{dist}
\end{figure}

\begin{table}[t]
\begin{center}
\renewcommand{\arraystretch}{2}
\resizebox{1\linewidth}{!}{
\begin{tabular}{l|cccc}
\toprule
{} & \textbf{Relevance} & \textbf{Coherence} & \textbf{Fluency} & \textbf{Helpfulness}\\
\midrule
Average Score & 4.37 & 4.31 & 4.46 & 4.20  \\

\bottomrule
\end{tabular}
}
\caption{User study results based on a 5-point Likert scale, evaluating four aspects: Relevance, Coherence, Fluency, and Helpfulness.}
\label{tab:user-study}
\end{center}
\end{table}

To evaluate the performance of our dialogue generation system, we conduct a user study involving five annotators, each with a background in machine learning and experience in developing conversational AI systems. Every annotator is tasked with reviewing the same set of 20 dialogues between a user and an AI assistant. These dialogues cover a variety of tasks, including product platform operations, troubleshooting issues, and general activities such as booking a restaurant or hotel.

In the user study, annotators are asked to rate each dialogue across four aspects: relevance, coherence, fluency, and helpfulness, defined as follows. 

\begin{itemize}[leftmargin=10pt]
    \item \textbf{Relevance}: Assesses how accurately the \name{} system's responses align with and address the user's queries.
    \item \textbf{Coherence}: Evaluates the logical flow and consistency of the \name{} system's responses throughout the conversation. Specifically, it measures how well each reply is contextually appropriate, connected to previous exchanges, and aligned with the user's goal, without introducing contradictions or abrupt shifts in topic.
    \item \textbf{Fluency}: Evaluates the grammatical correctness and naturalness of the \name{} system's language.
    \item \textbf{Helpfulness}: Measures the overall helpfulness of the \name{} system that enables the user to complete their task.
\end{itemize}

Among these aspects, Relevance and Fluency focus on individual interaction turns, while Coherence and Helpfulness are assessed based on the entire dialogue. Detailed descriptions of these four aspects provided to the annotators can be found in Appendix~\ref{app:metrics}. Each aspect is rated on a 5-point Likert scale, where 1 indicates the lowest score (e.g., not relevant at all) and 5 indicates the highest score (e.g., completely relevant). The annotation interface allows the annotators to view a screenshot of the conversation and rate each aspect, with optional comment sections to provide further insights or clarify their ratings.

The main results are shown on Table~\ref{tab:user-study}, which includes the average evaluation scores from five annotators. The results show that \name{} performs well across all four aspects, with average scores above 4.
The detailed distribution of evaluation scores, shown in Figure~\ref{dist}, provides further insight into the system's performance. Most of the scores are concentrated at 4 or 5. However, for Coherence and Helpfulness, there is a noticeable presence of scores at 2 or 3. Since both aspects assess the quality of the entire dialogue, this suggests that maintaining consistent performance throughout a conversation remains a challenge, particularly as the conversation length increases.

\section{Case Study}

\begin{figure*}[ht!]
  \centering
  \includegraphics[width=\linewidth]{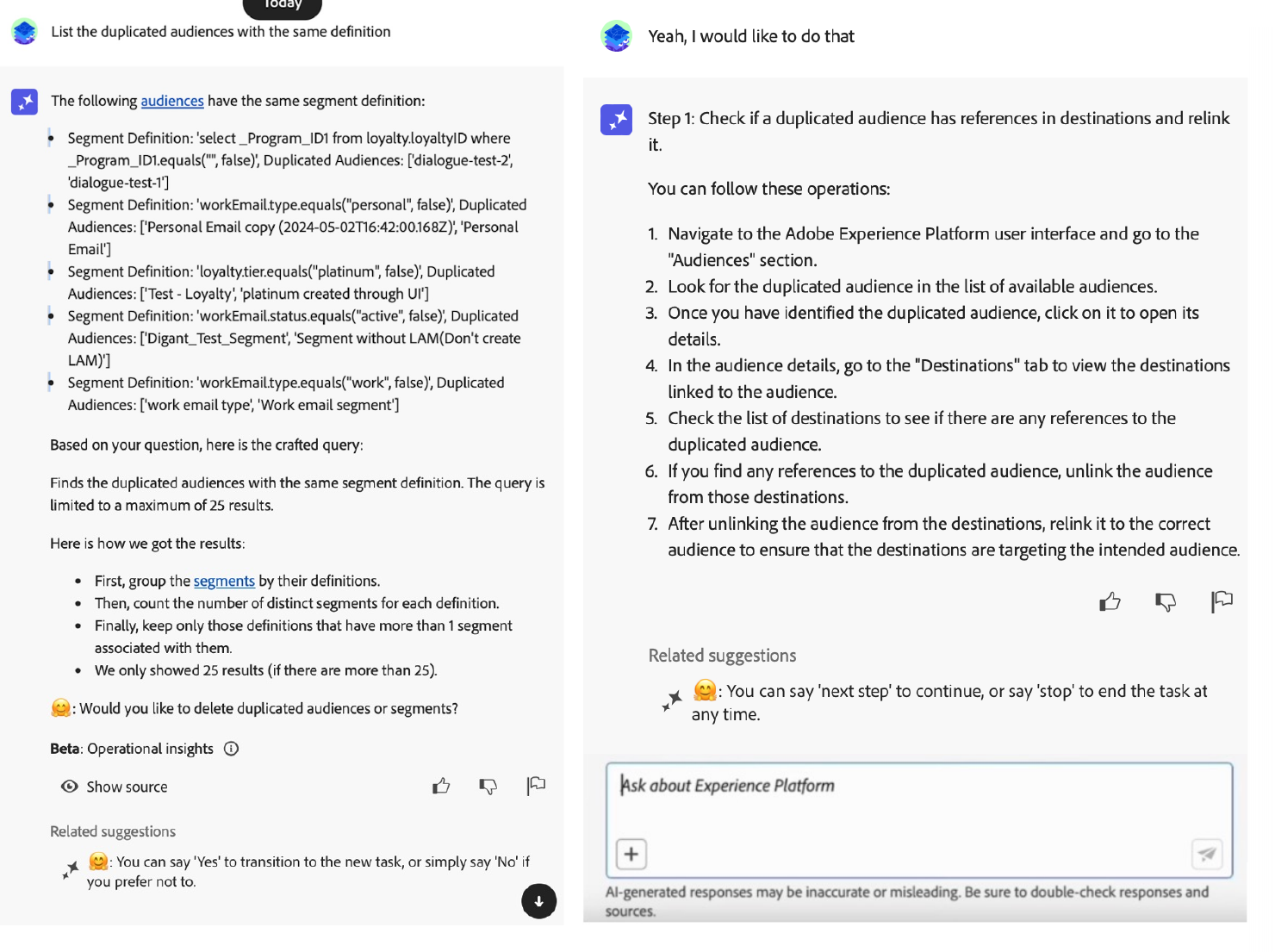}
  \caption{Case study. The first two turns of an example dialogue in which the user asks to list duplicated audiences.}
  \label{case_study}
\end{figure*}

We conduct a case study in this section to better understand the strengths and limitations of the system. As illustrated in the dialogue example in Figure~\ref{case_study}, the user initially asks to list duplicated audiences with the same segment definition. Our \name{} system successfully executes this request and returns the duplicated audiences. Moreover, it identifies a related sub-goal in the goal repository and transitions to a broader goal to delete the duplicated audiences, actively asking the user whether they would like to proceed. This reflects the system’s ability to think one step ahead, making it more proactive and intelligent.

However, some users in the study rated the system low on helpfulness in this case, providing feedback that for certain steps, such as relinking a duplicated audience from the destinations, it would be more useful if the system directly executed the action rather than offering step-by-step instructions. While our system currently supports some API integrations, such as listing duplicated audiences, direct execution of internal actions that involve sensitive knowledge or customer data still requires significant effort in fully integrating enterprise-specific toolkits and internal APIs. Nonetheless, this points toward a promising future direction: evolving \name{} into a more agentic system that can access internal APIs, perform web search, retrieve documentation, and ground its actions in outputs that better fulfill user needs. The interface of our system is provided in Appendix~\ref{app:interface}, along with more dialogue examples that cover both dialogue paradigms, including step-by-step guidance for task execution and slot-filling for information collection.


\section{Conclusion}

In this work, we introduce \name{}, a modular task-oriented dialogue system designed to assists users in completing tasks within enterprise environments. Our system features a comprehensive set of modules, and adopts a goal-driven approach to dialogues, making it more proactive and enabling mixed-initiative interactions. Furthermore, we implement a unified framework that integrates two representative TOD paradigms including slot-filling for information collection and step-by-step guidance for task execution. Results from our human study confirm the effectiveness and helpfulness of our dialogue system.

\section*{Ethics Statement}

Privacy and safety are top priorities in our dialogue system. We ensure that no personal or identifying information is incorporated into conversations. To further protect users, the system includes a series of safety checks, proactively rejecting inappropriate task requests and preventing the display of potentially harmful tasks that could pose risks to users or their belongings.

\section*{Acknowledgments}

We would like to express our gratitude to the team at Adobe for their valuable comments and feedback: Anirudh Sureshan, Pawan Sevak, Jake Bender, Vaishnavi Muppala, Shaista Hussain, Shreya Anantha Raman, Rachel Hanessian, Danny Miller, and Pritom Baruah.


\bibliographystyle{ACM-Reference-Format}
\bibliography{custom}

\clearpage
\appendix

\section{Prompt Design}
\label{app:prompt}

\subsection{Prompt for Dialogue State Tracking}
\label{app:prompt_dst}

\begin{multicols}{2}
\begin{promptbox}
Capture entity values from last utterance of the conversation according to examples.
Capture pair "entity:value" separated by colon and no spaces in between. 
Format the output in JSON.
If not specified, leave the value empty. Values that should be captured are:
{slots}

### Here is the conversation between user and ai-assistant:
{chat_history}
<<user>>: {current_utterance}

Capture all the entity values based on the conversation above and format the output in JSON:
\end{promptbox}
\end{multicols}

\subsection{Prompt for Response Generation}
\label{app:prompt_rg}

\begin{multicols}{2}
\begin{promptbox}
You are a task-oriented dialogue system designed to assist users in completing specific tasks such as booking a hotel or booking a flight. Your goal is to gather all necessary information (slots) required to complete the task through a series of user interactions. If all required slots are collected, you should confirm that the task has been completed.

### Task:
{task}

### Filled Slots: 
{filled_slots}

### Missing Slots: 
{missing_slots}

### Here is the conversation between user and ai-assistant:
{chat_history}
<<user>>: {current_utterance}

### Requirements:
1. If there are remaining slots that need to be filled, generate a polite and contextually appropriate utterance to request the next missing piece of information from the user. Ask one missing slot at a time.
2. If all required slots have been filled, briefly summarize all the collected slot information without asking the user any questions.
3. If the user asks a question, exactly start the placeholder <ANSWER> as the response, followed by a polite and contextually appropriate utterance to request the next missing piece of information from the user.

Generate an contextually appropriate user-facing utterance based on the current task, slot information and the conversation. The generated utterance should be friendly, polite, and positive.

<<ai-assistant>>: 
\end{promptbox}
\end{multicols}

\clearpage
\subsection{Prompt for NL2Goal Translator}
\label{app:prompt_translator}

\begin{multicols}{2}
\begin{promptbox}
Given a paragraph that describes a specific goal and its associated workflow, your task is to generate a YAML snippet that structures the information into a list of steps. Each step should include a "name" field summarizing the step and a "description" field for explaining additional details. Ensure that the step numbers in the YAML snippet are consistent with the numbers in the workflow.

### Example:

I have a goal to resolve an issue where the data pipeline is failing at the transformation stage. The workflow to address this involves the following steps: first, investigate the transformation logic to ensure all mappings and transformations are correct; second, verify that the data sources are providing complete and accurate data; and third, check the pipeline logs for any errors that might indicate issues during the transformation process.

The corresponding YAML should look like this:

workflow:
    - goal: "Resolve the issue where the data pipeline is failing at the transformation stage."
    steps:
        - name: "Investigate the transformation logic."
        description: "Ensure that all mappings and transformations are correct."

        - name: "Data verification."
        description: "Verify that the data sources are providing complete and accurate data."

        - name: "Check for errors."
        description: "Look for any errors in the pipeline logs that indicate issues during transformation."

### Task:

{new_goal}

Generate the corresponding YAML snippet:
\end{promptbox}
\end{multicols}

\clearpage
\section{User Study Metrics}
\label{app:metrics}

In our user study, annotators are asked to rate each dialogue on a 5-point Likert scale across four aspects: relevance, coherence, fluency, and helpfulness. The detailed scoring criteria for all four aspects, including Relevance, Coherence, Fluency, and Helpfulness, are provided in the Table~\ref{relevance}--\ref{helpfulness}.





\begin{table}[h!]
\centering
\small
\renewcommand{\arraystretch}{1.5}
\begin{tabular}{c|m{5.5cm}}
\hline
\textbf{Score} & \centering \textbf{Relevance} \tabularnewline
\hline
1 & The AI assistant's responses are not relevant at all to the user's queries. \\
\hline
2 & The AI assistant's responses are mostly irrelevant to the user's queries. \\
\hline
3 & The AI assistant's responses are moderately relevant to the user's queries. \\
\hline
4 & The AI assistant's responses are mostly relevant to the user's queries. \\
\hline
5 & The AI assistant's responses are all relevant to the user's queries. \\
\hline
\end{tabular}
\caption{Detailed scoring criteria for Relevance in the user study.}
\label{relevance}
\end{table}

\begin{table}[h!]
\centering
\small
\renewcommand{\arraystretch}{1.5}
\begin{tabular}{c|m{5.5cm}}
\hline
\textbf{Score} & \centering \textbf{Coherence} \tabularnewline
\hline
1 & The AI assistant's responses throughout the conversation are not coherent at all. \\
\hline
2 & The AI assistant's responses throughout the conversation are mostly not coherent. \\
\hline
3 & The AI assistant's responses throughout the conversation are moderately coherent. \\
\hline
4 & The AI assistant's responses throughout the conversation are mostly coherent. \\
\hline
5 & The AI assistant's responses throughout the conversation are completely coherent. \\
\hline
\end{tabular}
\caption{Detailed scoring criteria for Coherence in the user study.}
\label{coherence}
\end{table}

\begin{table}[h!]
\centering
\small
\renewcommand{\arraystretch}{1.5}
\begin{tabular}{c|m{5.5cm}}
\hline
\textbf{Score} & \centering \textbf{Fluency} \tabularnewline
\hline
1 & The AI assistant's responses are not fluent at all. \\
\hline
2 & The AI assistant's responses are mostly not fluent. \\
\hline
3 & The AI assistant's responses are moderately fluent \\
\hline
4 & The AI assistant's responses are mostly fluent. \\
\hline
5 & The AI assistant's responses are all fluent. \\
\hline
\end{tabular}
\caption{Detailed scoring criteria for Fluency in the user study.}
\label{fluency}
\end{table}

\begin{table}[h!]
\centering
\small
\renewcommand{\arraystretch}{1.5}
\begin{tabular}{c|m{5.5cm}}
\hline
\textbf{Score} & \centering \textbf{Helpfulness} \tabularnewline
\hline
1 & The AI assistant's responses throughout the conversation are not helpful at all. \\
\hline
2 & The AI assistant's responses throughout the conversation are mostly not helpful. \\
\hline
3 & The AI assistant's responses throughout the conversation are moderately helpful. \\
\hline
4 & The AI assistant's responses throughout the conversation are mostly helpful. \\
\hline
5 & The AI assistant's responses throughout the conversation are completely helpful. \\
\hline
\end{tabular}
\caption{Detailed scoring criteria for Helpfulness in the user study.}
\label{helpfulness}
\end{table}

\section{Interface}
\label{app:interface}

\begin{figure*}[ht]
  \centering
  \includegraphics[width=\linewidth]{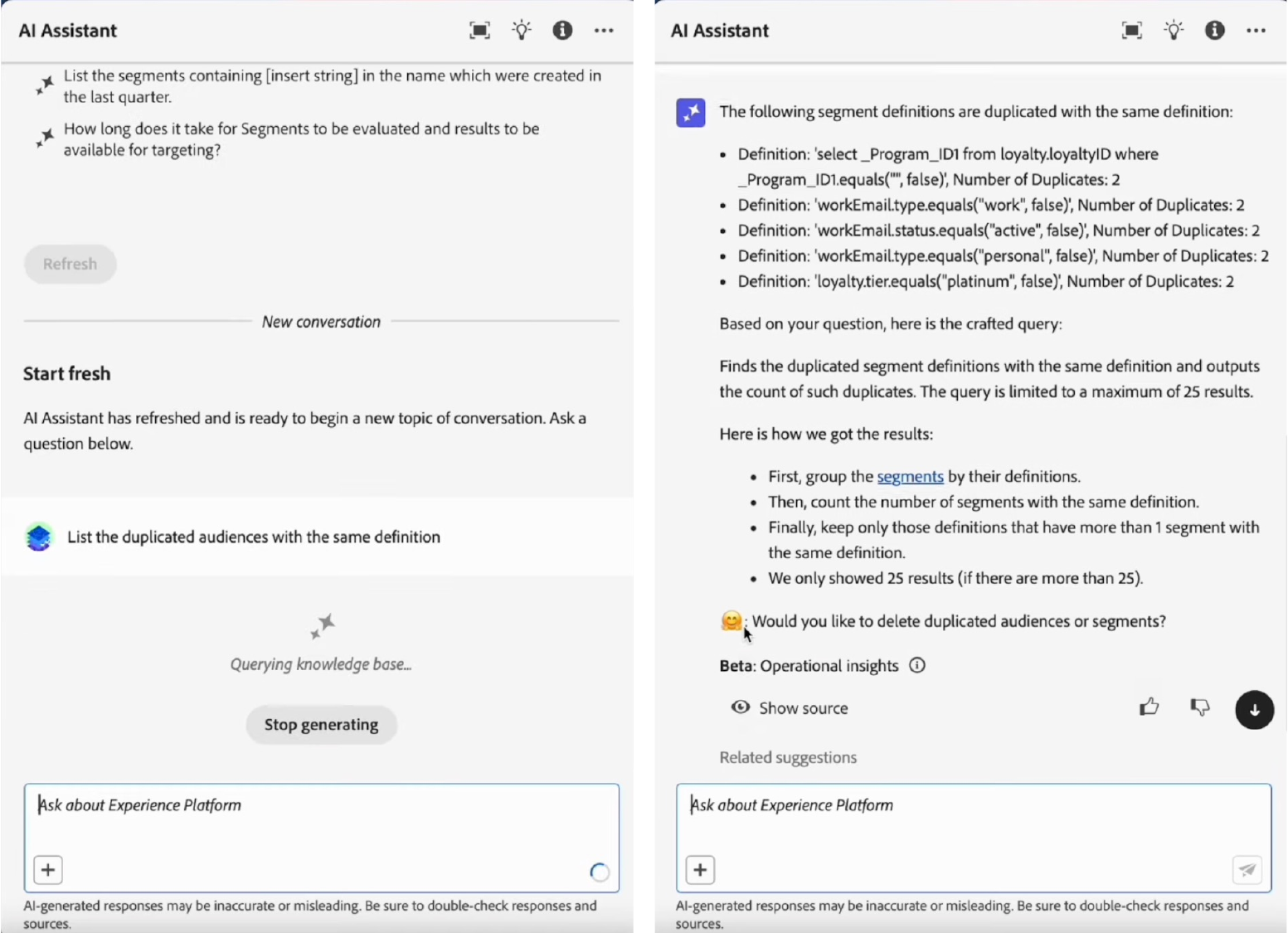}
  \caption{Dialogue A. It follows the step-by-step guidance paradigm for task execution. The example shown focuses on performing data hygiene for audience segments.}
  \label{dialogue1}
\end{figure*}

\begin{figure*}[ht]
  \centering
  \includegraphics[width=\linewidth]{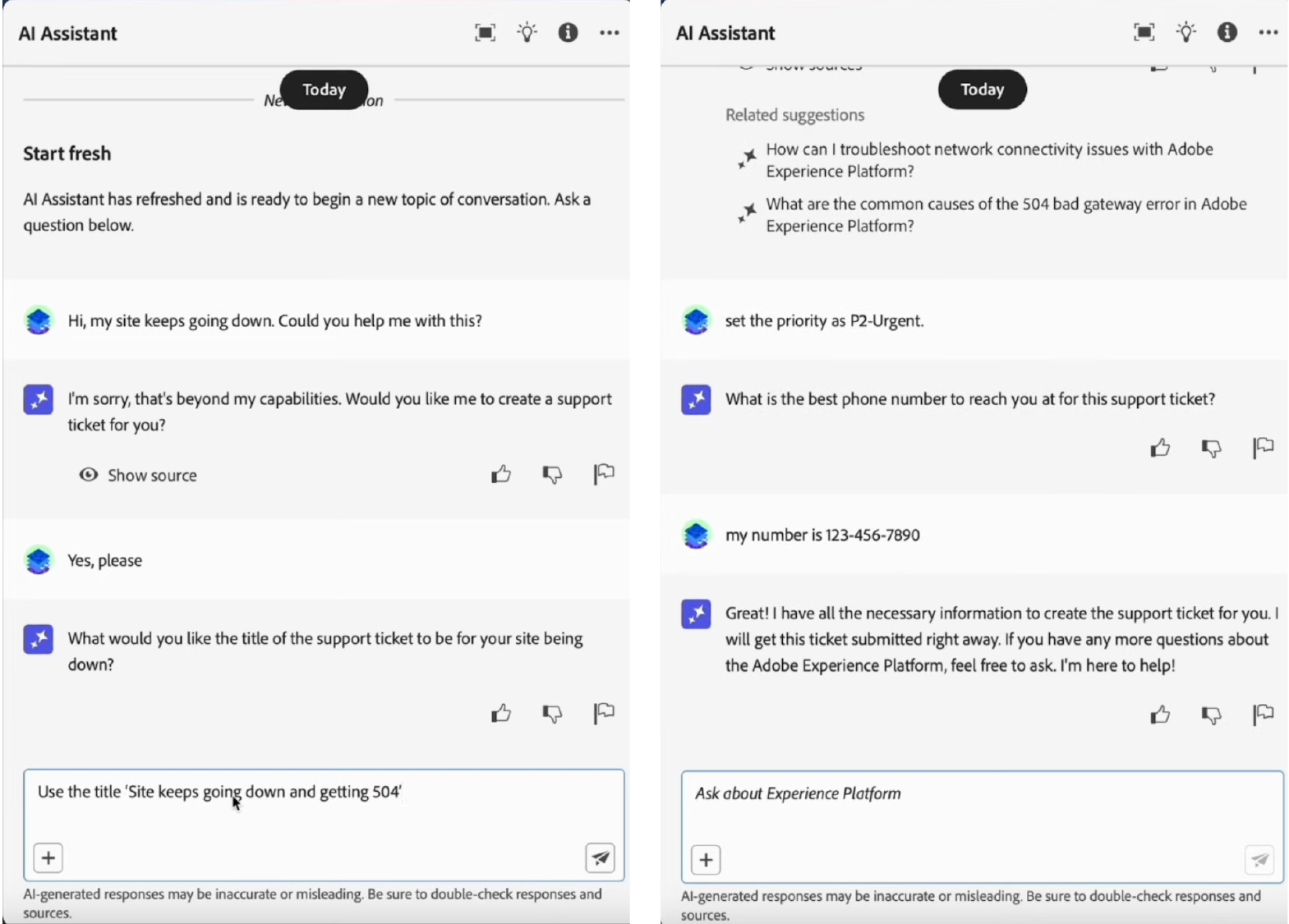}
  \caption{Dialogue B. It follows the slot-filling for information collection paradigm. The example shows the process of helping the user create a support ticket.}
  \label{dialogue2}
\end{figure*}

This section presents the user interfaces of the \name{} dialogue system, prototyped within the Adobe Experience Platform. Figures~\ref{dialogue1} and~\ref{dialogue2} show interface screenshots illustrating partial examples of the system’s two dialogue paradigms: step-by-step guidance for task execution and slot-filling for information collection. Full dialogue examples are available in the demo\footnote{Demo: \href{https://www.youtube.com/shorts/_v8f3x5IKEw}{https://www.youtube.com/shorts/\_v8f3x5IKEw}}.
 
\end{document}